\documentclass[10pt,twocolumn,letterpaper]{article}

\usepackage{cvpr}              %

\usepackage{multirow}
\usepackage{graphicx}
\usepackage{amsmath}
\usepackage{amssymb}
\usepackage{booktabs}
\usepackage{arydshln}

\usepackage[table,xcdraw]{xcolor}
\newcommand{\bgl}{\cellcolor[HTML]{DDDDDD}}
\newcommand{\bgd}{\cellcolor[HTML]{BBBBBB}}

\renewcommand{\arraystretch}{1.1} %

\newcommand{\ms}[1]{\textcolor{blue}{#1}}

\renewcommand{\thefootnote}{\fnsymbol{footnote}}

\usepackage[pagebackref,breaklinks,colorlinks]{hyperref}

\usepackage[capitalize]{cleveref}
\crefname{section}{Sec.}{Secs.}
\Crefname{section}{Section}{Sections}
\Crefname{table}{Table}{Tables}
\crefname{table}{Tab.}{Tabs.}

\usepackage[numbers,compress]{natbib}
\usepackage{siunitx}
\usepackage{mathtools}
\usepackage{ifthen}

\newcommand{\RP}{\textbf{RP}}

\newcommand{\N}{\mathbb{N}}

\newcommand{\matrixgroup}[2]{%
  \ifthenelse{\equal{#2}{}}
  {\ensuremath{\mathbf{#1}}}
  {\ensuremath{\mathbf{#1}(#2)}}}

\renewcommand{\l}{\ell}

\def\cvprPaperID{7077} %
\def\confName{CVPR}
\def\confYear{2022}

\begin{document}

\title{Light Field Neural  Rendering}

\author{
\begin{tabular}{cccc}
Mohammed Suhail$^{1,2}$\footnote[1]{Work done while interning at Google.} & Carlos Esteves$^{4}$ & Leonid Sigal$^{1,2,3}$ & Ameesh Makadia$^{4}$\\
 {\tt\small suhail33@cs.ubc.ca} & {\tt\small machc@google.com} &
 {\tt\small lsigal@cs.ubc.ca} & {\tt\small makadia@google.com}
\end{tabular}
\\
\begin{tabular}{lccr}
$^1$University of British Columbia &
$^2$Vector Institute for AI &
$^3$Canada CIFAR AI Chair &
$^4$Google
\end{tabular}
}

\maketitle

\begin{abstract}

Classical light field rendering for novel view synthesis
can accurately reproduce view-dependent effects such as reflection, refraction, and translucency,
but requires a dense view sampling of the scene.
Methods based on geometric reconstruction need only sparse views,
but cannot accurately model non-Lambertian effects.
We introduce a model that combines the strengths and mitigates the limitations of these two directions.
By operating on a four-dimensional representation of the light field,
our model learns to represent view-dependent effects accurately.
By enforcing geometric constraints during training and inference,
the scene geometry is implicitly learned from a sparse set of views. 
Concretely, we introduce a two-stage transformer-based model that first
aggregates features along epipolar lines,
then aggregates features along reference views
to produce the color of a target ray.
Our model outperforms the state-of-the-art on multiple forward-facing and 360$^\circ$ datasets,
with larger margins on scenes with severe view-dependent variations. Code and results can be found at \href{https://light-field-neural-rendering.github.io/}{light-field-neural-rendering.github.io}.
\let\thefootnote\relax\footnote{* Work done while interning at Google.}
\end{abstract}

\section{Introduction}
\label{sec:intro}
 
\begin{figure}
    \centering
    \includegraphics[width=.47\textwidth]{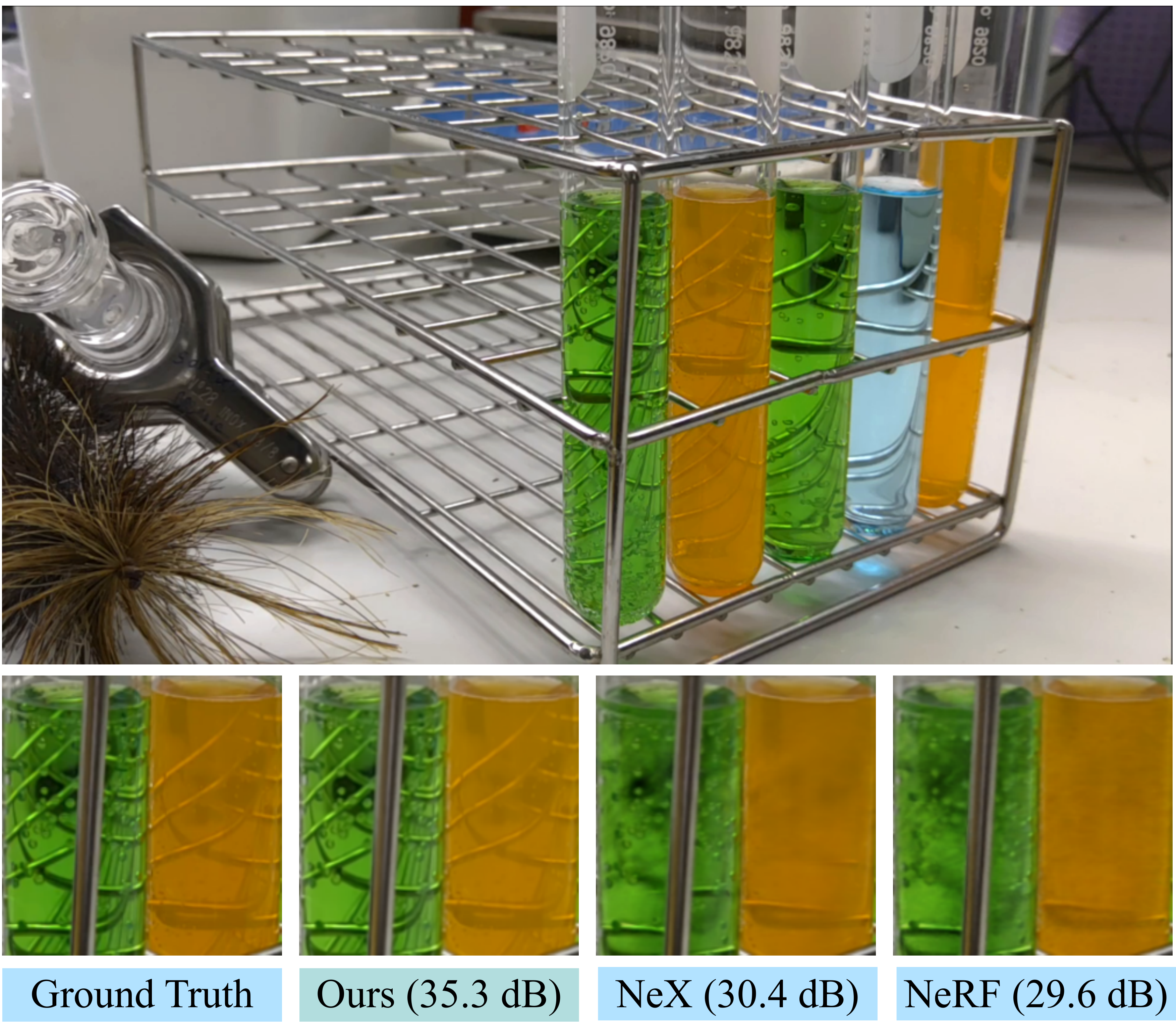}
    \caption{\textbf{Novel view synthesis.} On top is the target image to be rendered, from the \emph{Lab} scene in the Shiny dataset \cite{wizadwongsa2021nex}. Bottom row shows crops of novel views generated by our proposed model, NeX \cite{wizadwongsa2021nex},  and NeRF \cite{mildenhall2020nerf}. Unlike NeX and NeRF that fail to synthesize refractions on the test tube, our model almost perfectly reconstructs these complex view-dependent effects.
      We indicate the PSNR of the rendered images within parenthesis (higher is better).
      Images can be zoomed for detail.}
    \label{fig:intro_figure}
\end{figure}

Synthesizing a novel view given a sparse set of images is a long-standing challenge in computer vision and graphics \cite{chen1993view,shade1998layered,shum2000review}.   Recent advances in 3D neural rendering for view synthesis, in particular NeRF~\cite{mildenhall2020nerf} and its successors \cite{pumarola2021d,gafni2021dynamic,xian2021space,du2021neural,lindell2021autoint,oechsle2021unisurf}, have brought us tantalizingly close to the capability of creating photo-realistic images in complex environments.  One reason for NeRF’s success is its implicit 5D scene representation which maps a 3D scene point and 2D viewing direction to opacity and color. In principle, such a representation could be perfectly suited to modeling view-dependent effects such as the non-Lambertian reflectance of specular and translucent surfaces.  However, without regularization, this formulation permits degenerate solutions due to the inherent ambiguity between 3D surface and radiance, where an incorrect shape (opacity) can be coupled with a high-frequency radiance function to minimize the optimization objective~\cite{zhang2020nerf++}.  In practice, NeRF avoids such degenerate solutions through its neural architecture design, where the viewing direction is introduced only in the last layers of the MLP, thereby limiting the expressivity of the radiance function, which effectively translates to a smooth BRDF prior~\cite{zhang2020nerf++}. Thus, NeRF manages to avoid degenerate solutions at the expense of fidelity in non-Lambertian effects (\cref{fig:intro_figure} highlights this particular limitation of the NeRF model).  Photo-realistic synthesis of non-Lambertian effects is one of the few remaining hurdles for neural rendering techniques.

In this paper, we formulate view synthesis as rendering a sparsely observed light field.  
The 4D light field~\cite{levoy1996light}, which measures the radiance along rays in empty space, is often used for view synthesis~\cite{levoy1996light,buehler2001unstructured,kalantari2016learning}. Rendering a novel view from a densely sampled light field can be achieved with signal processing techniques (\eg, interpolation) and without any model of the 3D geometry, but no such straightforward method exists with sparse light fields. From sparse images, rendering often utilizes additional 3D geometric constraints, such as predicted depth maps~\cite{srinivasan2017learning,kalantari2016learning}, but performance is sensitive to accurate depth estimates which are difficult to obtain for non-Lambertian surfaces.  

Motivated by these limitations, we introduce a novel method for rendering a sparse light field. 
Our neural rendering function operates in the style of image based rendering, where a target ray is synthesized using only observed rays from nearby views. In lieu of explicit 3D information, our transformer based rendering function is trained to fuse rays from nearby views exploiting an additional inductive bias in the form of a multi-view geometric constraint, namely the epipolar geometry. 
As shown in \cref{fig:intro_figure}, our model is able to faithfully reconstruct the sharp details and lighting effect in the most challenging scene in the Shiny dataset \cite{wizadwongsa2021nex}.

\vspace{0.1in}
\noindent
\textbf{Contributions.} Our main contribution is the novel light field based neural view synthesis model, capable of photorealistic modeling of non-Lambertian effects (\eg, specularities and translucency). To address the core challenge of sparsity of initial views, we leverage an inductive bias in the form of a multi-view geometric constraint, namely the epipolar geometry, and a transformer-based ray fusion. The resulting model produces higher fidelity renderings for forward-facing as well as \ang{360} captures, compared to state-of-the-art, achieving up to $5$ dB improvement in the most challenging scenes. Further, as a byproduct of our design, we can easily obtain dense correspondences and depth without further modifications, as well as transparent visualization of the rendering process itself. Through ablations we illustrate the importance of our individual design choices.

\section{Related work}
\label{sec:relatedwork}

\noindent\textbf{Light field rendering.}
\citet{levoy1996light} defined the 4D light field as a function that specifies the radiance of any given ray in free space. 
They forwent geometric reasoning to directly synthesize novel views from input samples. Lumigraph rendering \cite{gortler1996lumigraph} exploits proxy geometry to counter aliasing effects that stem from irregularity or view under-sampling. Recent works \cite{kalantari2016learning,srinivasan2017learning,sitzmann2021light,buehler2001unstructured,wu2021revisiting} have explored learning based methods to light field rendering. These methods, however, either require dense input sampling \cite{kalantari2016learning}, have limited range of motion \cite{srinivasan2017learning} or are limited to simple scenes \cite{sitzmann2021light}.
In this work, we focus on novel view synthesis for
complex scenes with challenging non-Lambertian effects from a sparse set of viewpoints.

\vspace{.3em}
\noindent\textbf{Neural scene representation.}
Representing shape and appearance of scenes using neural networks has recently gained immense popularity. {\em Explicit} representation-based methods use differentiable rendering to learn 3D representation such as point clouds~\cite{wu2020multi,aliev2020neural,ruckert2021adop}, meshes~\cite{thies2019deferred} or voxels~\cite{sitzmann2019deepvoxels,lombardi2019neural} for the scene. {\em Implicit} representation-based methods represent scenes using continuous coordinate-based functions such as signed distance fields~\cite{atzmon2019controlling,jiang2020local,genova2020local,xu2019disn,chabra2020deep,yariv2021volume} or occupancy fields~\cite{mescheder2019occupancy,peng2020convolutional}. Scene Representation Networks~\cite{sitzmann2019srns} use a differentiable ray marching algorithm along with a continuous function that maps coordinates to features. NeRF~\cite{mildenhall2020nerf} achieves photo-realistic rendering by learning a function that maps points along a ray to color and opacity followed by volumetric rendering. NeX~\cite{wizadwongsa2021nex} is a multiplane image-based scene representation that addresses NeRF's difficulty to model large view dependent effects. However, NeX is still challenged in scenarios such as interference patterns caused by reflection or refraction through liquid. In this work, we introduce a model that can faithfully render novel views in the presence of complex view-based effects in scenarios where other methods fail (\eg, \cref{fig:intro_figure}). For a comprehensive survey of recent advances in neural rendering please refer to \citet{tewari2021advances}.

\vspace{.3em}
\noindent\textbf{Image-based rendering.}
Image-based rendering (IBR) methods \cite{debevec1996modeling,du2018montage4d,wang2021ibrnet,chaurasia2013depth} are built on the notions that novel views can be render by ``borrowing'' pixel values from a given set of input images. Global geometry-based methods such as \citet{riegler2021stable,hedman2018instant,riegler2020free,hedman2017casual} rely on dense reconstruction from input views to obtain a global mesh for the scene. These meshes are used for projecting target rays onto nearby view for feature and color extraction. Other methods such as \citet{penner2017soft,chaurasia2013depth} infer depth maps using multi-view stereo methods to compute warping transforms from the given input to target viewpoints. \citet{thies2018ignor} combine IBR with GAN-based image synthesis to learn view-dependent effects. To overcome difficulties caused by error in depth estimation, %
\citet{choi2019extreme} estimate a depth uncertainty distribution to refine images. 

Recently, \citet{wang2021ibrnet} introduced IBRNet, a NeRF-based model that incorporates features from nearby views for rendering. Their architecture predicts colors for each point on the ray as weighted average of colors from neighboring views. The densities for each point are predicted by aggregating information from all other points using a single attention layer. Similarly for category-specific reconstruction, NerFormer~\cite{reizenstein2021common} proposed to replace the MLP in NeRF-WCE~\cite{henzler2021unsupervised} with a transformer model to allow for spatial reasoning.
Our work crucially differs from these methods at their core: the rendering framework.
Our method employs a light field representation, forgoing the need for volumetric rendering. Furthermore, we introduce a transformer-based model that first reasons about correspondences to aggregate features along each epipolar line, then reasons about occlusion and lighting effects to aggregate features from multiple views to produce the final color.

\section{Approach}
\label{sec:approach}
\begin{figure*}[htp]
    \centering
    \includegraphics[width=\textwidth]{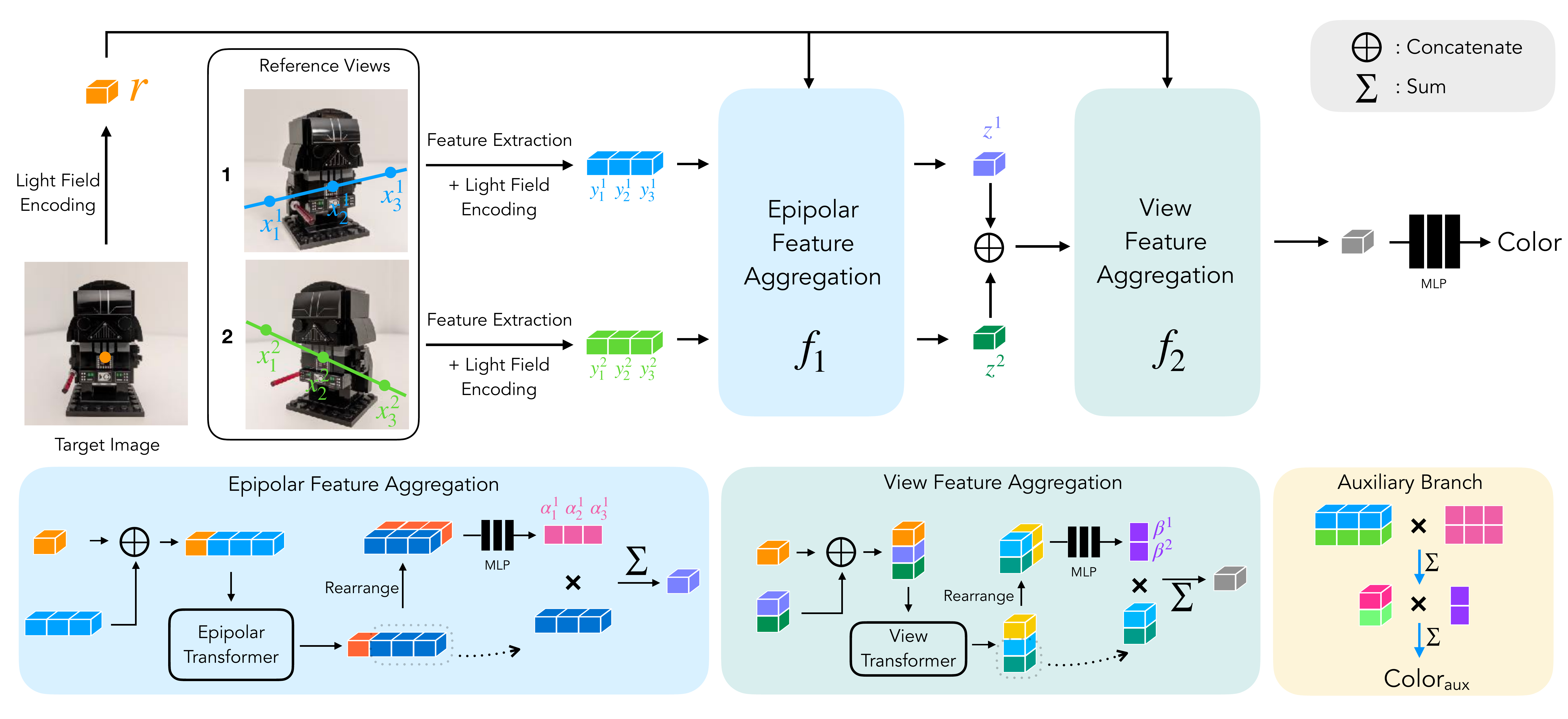}
    \caption{\textbf{Model Overview.} 
      Given a target ray to render,
      we identify reference views and sample points along the epipolar lines corresponding to the target ray.
      Features of these epipolar points along with the light field coordinates of the target ray
      are inputs to the epipolar aggregation.
      This stage (blue), \emph{independently} aggregates
      features along the epipolar lines for each reference view, producing reference view features.
      The reference view features along with the target ray are passed to the view aggregation stage (green),
      which combines the reference view features to predict the target ray color.
    }
    \label{fig:model_overview}
\end{figure*}
Our goal is to synthesize novel views of a scene given a collection of input images available during both training and inference.
Our design is guided by two key ideas,
1) using the four-dimensional parametrization of the light-field as input enables capturing view-dependent effects with high fidelity, and
2) enforcing constraints from multiple view geometry allows for view synthesis with sparse input views.

These ideas enable faithfully recovering illumination effects as in classical lightfield methods~\cite{levoy1996light,gortler1996lumigraph},
but require only a sparse view-sampling of the scene, as in geometry-based methods~\cite{debevec1996modeling,narayanan1998constructing} which traditionally struggle reproducing non-Lambertian effects.
To implement them, we introduce an epipolar-geometric inductive bias in conjunction with a transformer-based architecture.
Our model can render novel views from forward-facing photos as well as \ang{360} scenes captured with cameras on a hemisphere.

In the following sections, we first introduce the light field representation, then an overview of the model, followed by a detailed description of the network architecture.

\subsection{Light field parametrization}
\label{sec:lfparametrization}
Light fields are functions on the space of oriented lines that associate a radiance value to a given ray. In free space, as the radiance along the ray remains constant, the space of rays has four degrees of freedom and can be parametrized by 4D vectors.
We consider two distinct parametrizations of light field,
the light slab~\cite{levoy1996light} and the two-sphere~\cite{camahort1998uniformly}.
\vspace{-1.2em}
\paragraph{Light slab.} We adopt the light slab parametrization for forward-facing captures. A light slab consists of two parallel planes with their respective 2D coordinate systems  $(s, t)$ and $(u, v)$. Rays are then represented as a 4D tuple $r=(s, t, u, v)$ containing the coordinates of intersections with the two planes in their respective coordinate frames.
\vspace{-1.2em}
\paragraph{Two-sphere.} For \ang{360} scenes, we use the two-sphere parametrization~\cite{camahort1998uniformly} of the light field. Given a sphere bounding a scene, rays from the camera are represented using the colatitudes and longitudes at the two intersections with the sphere, $r=(\theta_1, \phi_1, \theta_2, \phi_2)$.

\vspace{0.06in}
Given a four-dimensional light field ray parametrization $r$, we learn a neural rendering model
$f$ that maps the rays to radiance values.

To obtain the ray coordinates for a given pixel in homogeneous coordinates $x \in \RP^2$,
from an image taken using a camera with intrinsics $C$ and pose (extrinsics) $[R\; t]$,
we first obtain the ray as a line $\ell$ in world coordinates parametrized by $\delta$ as
$\ell(\delta) = -R^\top t + \delta R^\top C^{-1}x$,
then solve for $\delta$ to obtain the intersections $r$ with either the two planes or the sphere.
To render an image, we evaluate the model $f(r)$ for the rays associated to each target pixel.

\subsection{Model overview}
Optimizing a neural rendering model $f$ that directly maps 4D light field coordinates to color fails to generalize to novel views when trained with a sparse set of input views (see \cref{subsec:ablation} for quantitative evaluation).

To address this challenge, we introduce a model
that incorporates a geometric inductive bias in the form of the epipolar constraints.

Given a target camera, we identify a set of neighboring views to be used to enforce multiple-view consistency.
During training, this set is constructed by randomly choosing $K$ views from a subset of $N$ closest views.
During inference, the closest $K$ are chosen deterministically.
We refer to the set of $K$ chosen views as \emph{reference views}.

Now given a target pixel $x$ to be rendered,
we obtain its ray parametrizations $\ell$ and $r$ as described in \cref{sec:lfparametrization},
sample a sequence of $P$ points $p_i = \ell(\delta_i)$ along the ray,
and project each point to each reference view as
$x_i^j = C_j[R_j\; t_j]p_i$,
where $C_j$, $[R_j\; t_j]$ are the reference view camera intrinsics and extrinsics, respectively,
and $1 \le j \le K$.

The collection $x^j = \{x_i^j\}_{1 \le i \le P}$ consists of points along the epipolar line of the target ray in the $j^{th}$ reference view.
and we refer to $x_i^j$ as \emph{epipolar points}.
To each epipolar point, we associate its ray parametrization as
described in \cref{sec:lfparametrization},
yielding the collections $r^j = \{r_i^j\}_{1 \le i \le P}$.
\vspace{-1.2em}
\paragraph{Epipolar feature aggregation.}
The first stage of our model, represented by the function $f_1$,
computes a feature representation per reference view by
aggregating features associated to the epipolar points and target ray.
We detail what those features are in the following sections.
Conceptually, the first stage computes the set of features
$\{z^j\}_{1 \le j \le K}$ where $z^j = f_1(r, r^j)$.
This is loosely related to classical multiple view geometry,
where we look for a correspondence to the target ray along the epipolar line.
In our case, however, there is no visual representation of the target ray,
so the model must learn to match the target ray coordinates
with the available reference features, and the output is a feature vector
representing the view $j$.
\vspace{-1.2em}
\paragraph{View feature aggregation.}
The second stage, represented by the function $f_2$,
predicts the target ray color by aggregating features associated to each reference view,
given the target ray representation.
Conceptually, the color for pixel $x$ with associated ray $r$ 
is predicted as $f(r) = f_2(r, \{f_1(r, r^j)\})$.
This stage learns to reason about occlusion and illumination effects to combine
information from all views and produce the target ray color.

\subsection{Network architectures}
\label{sec:transformer}

One possible approach would be to model $f_1$ and $f_2$ as multi-layer perceptrons (MLP).
However, this impedes the model from exploiting readily available relational information that can be extracted from the epipolar points, leading to sub-optimal performance (\cref{subsec:ablation} quantifies this).
Since inputs to $f_1$ are a \emph{sequence} of epipolar points,
and inputs to $f_2$ are a \emph{set} of reference view features,
we propose to use transformers,
which excel in sequence and set modeling, to model both epipolar and view feature aggregation.
\begin{figure*}[!t]
    \centering
    \includegraphics[width=.95\textwidth]{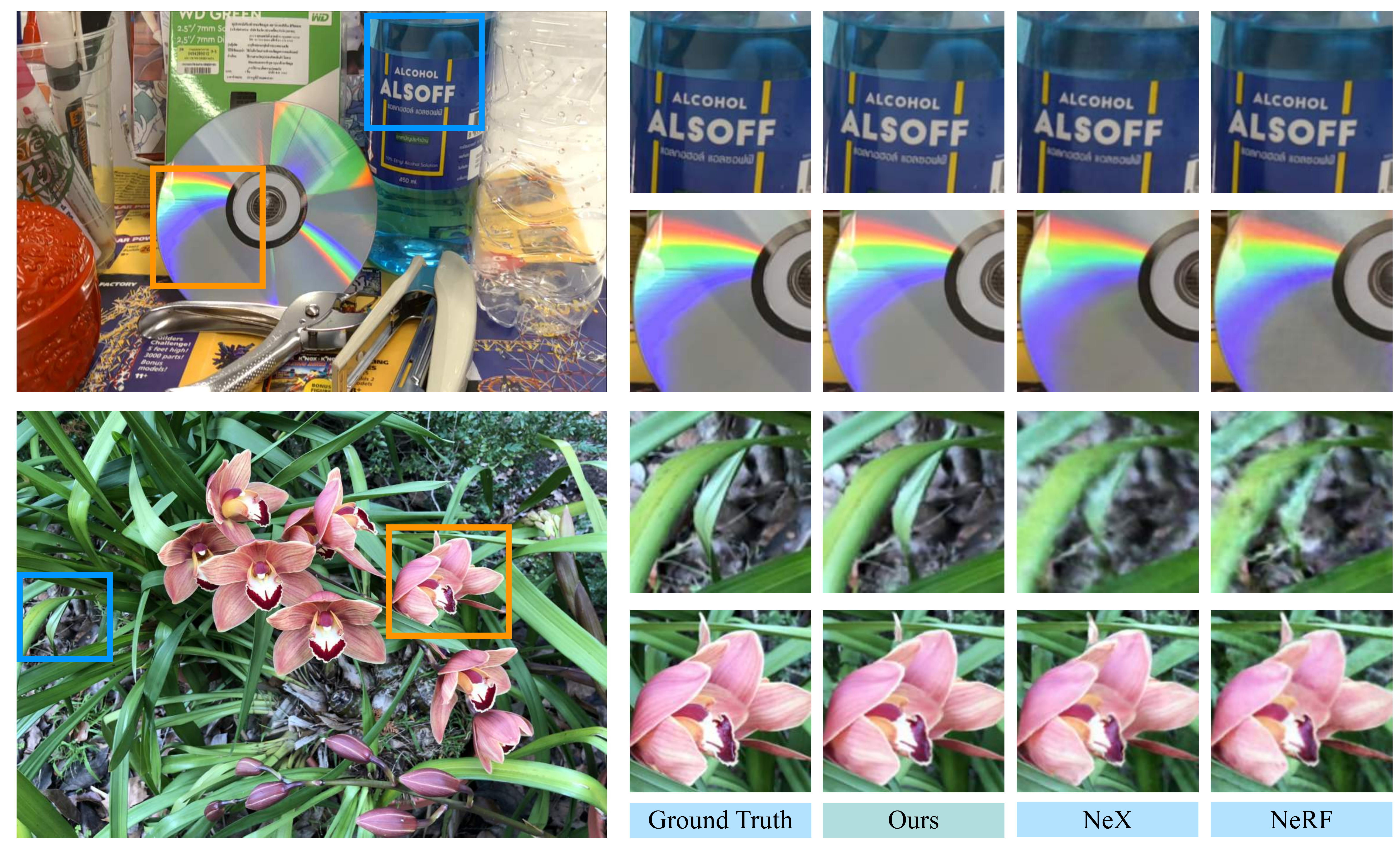}
    \caption{\textbf{Qualitative Comparison.}
      Top: results on the \emph{CD} scene from Shiny dataset~\cite{wizadwongsa2021nex}.
      Our method is able to retrieve sharper details in the reflections on the bottle
      (\eg, the top-left of the insets) 
      as well as the interference patterns on the compact disk
      (\eg, rainbow and reflection on the top-right).  
      Bottom: results on the \emph{Orchids} scene from the real-forward facing (RFF) dataset~\cite{mildenhall2019local}.
      Our method recovers more accurately the shape of the leaves.
      We also observe sharper texture on the leaves as well as the petals.}
    \label{fig:qualitative}
\end{figure*}
\subsubsection{Epipolar feature transformer ($f_1$)}
This transformer, highlighted in blue in \cref{fig:model_overview},
combines the features of points along the epipolar line based on the target rays.

The input is a sequence of $P+1$ features,
with $P$ features from epipolar points and one from the target ray.
The feature vector for the target ray is its own coordinates $r$.
The feature for an epipolar point $x_i^j$ is a concatenation of
1) ray coordinates $r_i^j$,
2) coordinates of $p_i$, the 3D point along $r$ projected to $x_i^j$,
3) a learnable camera embedding $k_j$,
4) visual features $v_i^j$ at $x_i^j$, obtained from a lightweight CNN, and
5) the color $c_i^j$ at $x_i^j$.

Assuming the target pixel $x$ matches to an epipolar point $x_i^j$, the corresponding point in the scene can be solved for and will have coordinates of $p_{i}$.
Including it as an epipolar point feature also plays the role
of positional encoding, since each point in the epipolar line correspond
to some depth value along the query ray.
This type of positional encoding is richer than the typical 1D encoding used
in sequence modeling~\cite{aiayn}, and more appropriate for modeling a 3D scene,
as demonstrated in \cref{subsec:ablation}.

We further apply Fourier features~\cite{mildenhall2020nerf,tancik2020fourfeat}
positional encoding to facilitate learning of high-frequency functions.
This operation is performed by $\gamma_r$ for ray coordinates and $\gamma_p$ for point coordinates,
see \cref{subsec:imp_details} for details.
To summarize, each epipolar point $x_i^j$ is represented by a feature
\begin{align}
  y_i^j = [\gamma_r(r_i^j) \parallel \gamma_p(p_i) \parallel k_j \parallel v_i^j \parallel c_i^j],
\end{align}
where $\parallel$ denotes concatenation.
The epipolar transformer for view $j$ will take as inputs $[\gamma_r(r), \{y_i^j\}_{1 \le i \le P}]$.
A linear layer first projects the features to the same dimension,
then a self-attention transformer is applied to the whole sequence.

We aggregate the $P$ outputs corresponding to the epipolar points ($\tilde{y}_i^j$)
to obtain the \emph{reference view features}.
The aggregation is a weighted average, with the weights computed using an attention mechanism similar to the Graph Attention Networks (GAT)~\cite{velickovic2018graph} as follows,
\begin{equation}
  \alpha_i^j =  \frac{
    \exp \left( W_1 \left[\tilde{r} \parallel {\tilde{y}_i^j} \right] \right)}
  {\sum_{k} \exp \left( W_{1} \left[ \tilde{r} \parallel {\tilde{y}_k^j} \right] \right)},
\label{eq:epipolar_avg}
\end{equation}
where $\tilde{r}$ are the output features of the target ray, and $W_{1}$ are learned weights.
The first stage is completed by repeating
$z^{j} = f_1(r, r^j) = \sum_{i=1}^{P} \alpha_{i}^j \tilde{y}_i^j$
for all views $1 \le j \le K$.

\subsubsection{View feature transformer ($f_2$)}

This transformer, highlighted in green in \cref{fig:model_overview},
takes the target ray and the set of features for each reference view.
The input sequence is now $[\gamma_r(r), \{z^{j}\}_{1 \le j \le K}]$,
where $z^j$ are the reference view features computed by the first stage,
and the output is a single feature vector for the target ray.
We use the same self-attention transformer architecture as the epipolar feature aggregator.
The transformer output sequence $[\hat{r}, \{\tilde{z}^{j}\}_{1 \le j \le K}]$
is aggregated with a weighted average using the same idea as the previous section.
We compute the weights $\beta^{j}$ with learnable weights $W_{2}$,
\begin{equation}
     \beta^{j} =  \frac{
     \exp \left( W_{2} \left[ {\hat{r}} \parallel {\tilde{z}^{j}} \right] \right)}
     {\sum_{k} \exp \left( W_{2} \left[ \hat{r} \parallel {\tilde{z}^{k}} \right] \right)},
\label{eq:neighbor_avg}
\end{equation}
then the output of this stage is the target ray feature 
$\sum_{k=1}^{K} \beta^{k} \tilde{z}^{k}$,
which is linearly projected and passed by a sigmoid to produce the pixel color prediction $c$.

\subsection{Loss} 
During training, we minimize the $\mathcal{L}_2$ loss between the observed and predicted colors.
We additionally include an auxiliary loss to encourage the attention weights for the epipolar points ($\alpha_i^j$) and reference views ($\beta^j$) to be interpretable,
in the sense that high values of $\alpha_i^j$ suggest a valid match to the target ray,
while low values of $\beta^j$ might indicate occlusion.
This auxiliary loss also leads to more accurate renderings (see \cref{subsec:ablation}).
To compute it, we use the attention weights
to combine reference pixel colors and make a second color prediction as
\begin{equation}
    c_{\text{aux}} = \sum_{j} \beta^{j} \left( \sum_{i} \alpha_{i}^j c_{i}^{j} \right),
    \label{eq:reg_color}
\end{equation}
where $c_{i}^{j}$ is the color of the epipolar point $x_i^j$.
The auxiliary loss is then defined as the $\mathcal{L}_2$ loss between $c_{\text{aux}}$
and the ground truth.
The effect of this loss is two-fold: 
1) it incentivizes weights $\alpha_{i}^{j}$ to have lower entropy to avoid blurry predictions in the auxiliary branch, and
2) it encourages weights $\beta^j$ to be high for unoccluded views.

\section{Experiments}
\label{sec:experiments}

We show quantitative and qualitative comparisons against state-of-the-art methods for novel view synthesis. We also perform an ablation study to analyze the effectiveness of the components  introduced in our method.

\begin{table}[!t]
\small
\centering
\begin{tabular}{@{}lcccc}
\toprule
\toprule
Model                           & PSNR [dB] $\uparrow$ & SSIM $\uparrow$ & LPIPS  $\downarrow$ & Avg.  $\downarrow$ \\ \midrule
LLFF \cite{mildenhall2019local} & 24.41                & 0.863            & 0.211                & 0.0656              \\
NeRF \cite{mildenhall2020nerf}  & 26.76                & 0.883            & 0.246                & 0.0562              \\
IBRNet \cite{wang2021ibrnet}    & 26.73                & 0.851            & \bgl 0.175           & 0.0523              \\
NeX  \cite{wizadwongsa2021nex}  & \bgl27.26            & \bgl0.904        & 0.178                & \bgl0.0473          \\
Ours                            & \bgd28.26            & \bgd0.920        & \bgd0.062            & \bgd0.0297          \\ \bottomrule \bottomrule
\end{tabular}
\caption{Results for the real forward-facing (RFF) dataset~\cite{mildenhall2019local}.}
\label{table:RFF}
\end{table}
\subsection{Implementation details}
\label{subsec:imp_details}

\paragraph{Network architecture.}
We use similar transformer architectures as the ones recently introduced for vision related tasks~\cite{dosovitskiy2020image}. Each block consists of a single-headed self-attention layer and an MLP with Gaussian error linear unit (GELU) activation~\cite{hendrycks2016gaussian}. A residual connection is applied at every block, followed by a LayerNorm (LN)~\cite{layernormBa2016}.
Each transformer has 8 blocks and the internal feature size is 256.
The visual features $v_i^j$ are produced by a single convolutional layer with \numproduct{5 x 5} filters
and 32 channels.
\vspace{-1.2em}
\paragraph{Positional encoding.}
Following prior work~\cite{mildenhall2020nerf,tancik2020fourfeat},
we use Fourier features to encode input coordinates
to facilitate learning the high-frequency components required for accurate rendering. For the light slab parametrization and the 3D points $p_i$, we positionally-encode each ray coordinate~\cite{mildenhall2020nerf} 
as $\gamma_r(w) = \gamma_p(w) = \{\sin(2^kw)\}\cup \{\cos(2^kw)\}$ for $0 \le k \le 4$.
For the two-sphere parametrization, we found it beneficial to use a positional
encoding based on evaluating the spherical harmonics at the points $(\theta_1, \phi_1)$
and $(\theta_2, \phi_2)$, see the appendix for details.
The learnable camera embeddings $k_j$ are 256-dimensional.

\vspace{-1.2em}
\paragraph{Training/inference details.}
In each training step, we randomly choose a target image and sample a batch of random rays from it. The batch sizes are 4096 for the forward-facing datasets and 8192 for Blender. We train for \num{250000}  iterations with the Adam optimizer~\cite{adamKingmaB14} and a linear learning rate decay schedule with \num{5000} warm-up steps.
For inference, we sample contiguous blocks of rays to make a batch.
Training on a Blender scene takes around 23 hours on a 32-core TPUv3 slice. 
Rendering an \numproduct{800 x 800} image then takes around \num{9.2} seconds.

\subsection{Results}
We compare our method with LLFF~\cite{mildenhall2019local}, NeRF~\cite{mildenhall2020nerf}, IBRNet~\cite{wang2021ibrnet}, NeX~\cite{wizadwongsa2021nex} and Mip-NeRF~\cite{barron2021mip}.
We compare against Mip-NeRF only on the Blender dataset
because for forward-facing captures, as noted by \citet[Appx D]{barron2021mip},
Mip-NeRF performs on par with NeRF.

\vspace{-1.2em}
\paragraph{Metrics.}
To measure the performance of our model we use three widely adopted metrics: peak signal-to-noise ratio (PSNR), structural similarity index measure (SSIM), and the learned perceptual image patch similarity (LPIPS)~\cite{zhang2018perceptual}. Following \cite{barron2021mip}, we additionally report the geometric mean of $10^{-\text{PSNR}/10}$, $\sqrt{1 - \text{SSIM}}$ and LPIPS, which provides a summary of three metrics for easier comparison. We report the averages of each metric over all the scenes in each dataset. Please refer to the appendix for a scene-wise breakdown of the results.

\begin{table}[!t]
\small
\centering
\begin{tabular}{@{}lcccc}
\toprule
\toprule
Model                                        & PSNR $\uparrow$ & SSIM $\uparrow$ & LPIPS  $\downarrow$ & Avg.  $\downarrow$ \\ \midrule
NeRF \cite{mildenhall2020nerf}               & 25.60           & 0.851            & 0.259               & 0.0651             \\
NeX  \cite{wizadwongsa2021nex}               & 26.45           & \bgl0.890        & 0.165               & 0.0499             \\
IBRNet\footnotemark[2] \cite{wang2021ibrnet} & \bgl26.50       & 0.863            & \bgl0.122           & \bgl0.0468         \\
Ours                                         & \bgd27.34       & \bgd0.907        & \bgd0.045           & \bgd0.0294         \\ \bottomrule \bottomrule
\end{tabular}
\caption{Results for the Shiny dataset from NeX~\cite{wizadwongsa2021nex}.}
\label{table:shiny}
\end{table}
\begin{table}[!t]
\small
\centering
\begin{tabular}{@{}lcccc}
\toprule \toprule
Model                                        & PSNR $\uparrow$ & SSIM $\uparrow$ & LPIPS  $\downarrow$ & Avg.  $\downarrow$ \\ \midrule
NeRF \cite{mildenhall2020nerf}               & 31.01           & 0.953           & 0.050               & 0.0194             \\
IBRNet \cite{wang2021ibrnet}                 & 28.14           & 0.942           & 0.072               & 0.0299             \\
Mip-NeRF \cite{barron2021mip}                & \bgl33.09       & \bgl0.961       & \bgl0.043           & \bgl0.0161         \\
Ours                                         & \bgd33.85       & \bgd0.981       & \bgd0.024           & \bgd0.0110         \\ \bottomrule \bottomrule
\end{tabular}
\caption{Results for the Blender dataset from NeRF~\cite{mildenhall2020nerf}.}
\label{table:blender}
\end{table}
\vspace{-.3em}
\subsubsection{Real-forward-facing (RFF) dataset}
The RFF dataset introduced by~\citet{mildenhall2019local} consists of $8$ forward facing captures of real-world scenes using a smartphone. 
For our experiments, we use the same resolution and train/test splits as NeRF~\cite{mildenhall2020nerf}.

\Cref{table:RFF} reports the average metrics across all $8$ scenes in the RFF dataset.
We show qualitative comparisons on the \emph{Orchids} scene in \cref{fig:qualitative}. Compared to the baselines, our method retains sharper detailed textures and produces consistent shape boundaries on the leaves and petals.

\vspace{-.3em}
\subsubsection{Shiny dataset}
The RFF dataset mostly consists of diffuse scenes with little view-dependent effects. The Shiny dataset introduced in NeX~\cite{wizadwongsa2021nex} presents 8 scenes with challenging view-dependent effects, captured by forward-facing cameras.
We use the same image resolution and splits as NeX.

We compare our model against NeX, IBRNet and NeRF on the Shiny dataset in \cref{table:shiny}. We report the average scores across all scenes in the dataset. Our model consistently improves over the state-of-the-art in all metrics.
We show qualitative analysis of rendering on a test view from the  \emph{CD} scene in \cref{fig:qualitative}. Our model is able to reconstruct the interference patterns on the disk and reflections on the bottle with higher level of detail as compared to baselines.
\footnotetext[2]{We fine tune the pretrained model available at  \url{https://github.com/googleinterns/IBRNet} on each scene in Shiny.}
{\renewcommand{\arraystretch}{.9}
\begin{table}[!t]
\small
\centering
\begin{tabular}{@{}lcccc}
\toprule \toprule
Model             & PSNR $\uparrow$ & SSIM $\uparrow$ & LPIPS  $\downarrow$ & Avg.  $\downarrow$ \\ \midrule
Vanilla-NLF       & 17.39           & 0.614            & 0.516                & 0.1802              \\ \midrule
$1$-MLP           & 21.33           & 0.774            & 0.208                & 0.0900              \\
$2$-MLP           & 26.16           & 0.896            & 0.076                & 0.0390              \\ \midrule
No CNN ($v_i^j$)  & 27.43           & 0.910            & 0.057                & 0.0314              \\
No 3D Coordinates & 28.17           & 0.920            & 0.047                & 0.0273              \\
No LCE ($k_j$)    & 28.23           & 0.926            & 0.045                & 0.0264              \\
Mean Pooling      & 28.39           & 0.929            & 0.043                & 0.0255              \\
No Auxiliary Loss & 28.43           & 0.931            & 0.043                & 0.0253              \\ \midrule
Ours              & 28.78           & 0.934            & 0.038                & 0.0235              \\ \bottomrule \bottomrule
\end{tabular}
\caption{Ablation study on the RFF dataset~\cite{mildenhall2019local},
 with \qty{25}{\%} the original resolution (\numproduct{504 x 378}).
  Refer to \cref{subsec:ablation} for details.}
\label{table:ablation}
\end{table}}

\subsubsection{Blender dataset}
Our model is capable of rendering novel views of \ang{360} scenes. To evaluate this case, we use the synthetic dataset introduced by~\citet{mildenhall2020nerf}.
Each scene consists of \numproduct{800 x 800} resolution images rendered from viewpoints randomly sampled on a hemisphere around the object. 

\Cref{table:blender} reports the average performance across all scenes in the Blender dataset.  Our model improves over NeRF, IBRNet and Mip-NeRF on all metric and achieves new state-of-the-art results.
On the materials scene, which contains reflections on metallic balls, we observe an improvement of around \qty{4}{\dB} on the PSNR metric when compared to Mip-NeRF. We present the full table along with qualitative comparisons in the appendix.

\subsection{Ablation studies}
\label{subsec:ablation}
To validate the effectiveness of different design decisions, we run the following ablation experiments.

\vspace{-1.2em}
\paragraph{Geometric inductive bias.} We train a model, called `Vanilla-NLF', that uses an MLP to predict the color of a ray given only its light field representation, without consideration of the scene geometry in form of epipolar constraints.

\begin{figure}[!t]
    \centering
     \includegraphics[width=.47\textwidth]{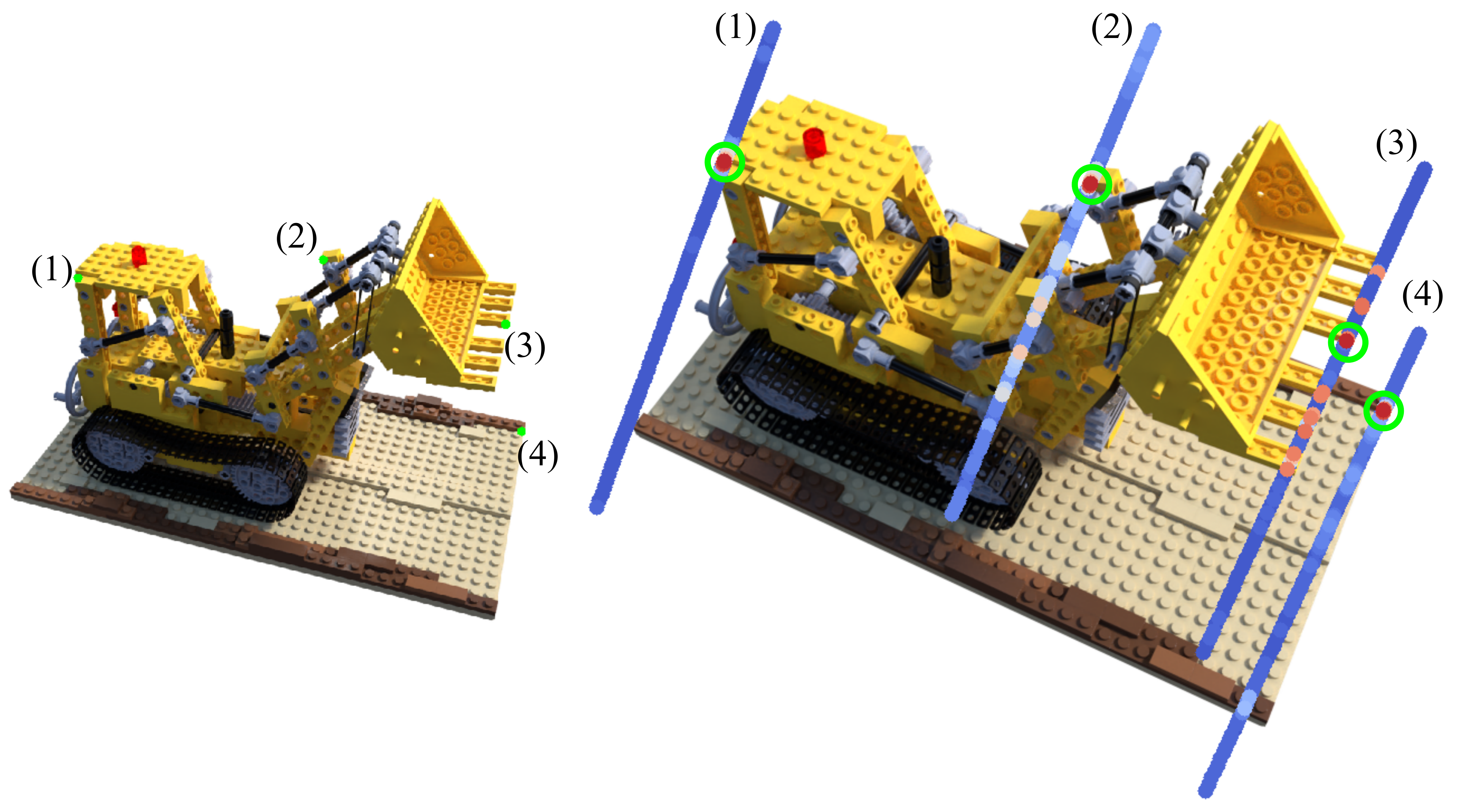} \\
     \caption{\textbf{Correspondence Distribution.}
       The per-point attention weights learned by our model indicate potential
       correspondences to the target ray. 
       We visualize four target rays and one reference view.
       The weights are in log-scale, from blue to red.
       The green circles highlight the point with highest correspondence probability.}
     
    \label{fig:correspondence}
\end{figure}

\vspace{.3em}
\noindent\textbf{Transformers vs MLPs.}
We train variations of our model replacing the transformers with MLPs. In the first variant, we replace each of the epipolar and view transformers with MLPs (`2-MLP'). The second variant replaces both transformers by a single MLP that takes as input all epipolar point features from all reference views along with the target ray and directly predicts the color (`1-MLP'). We detail the architectures in the appendix. We run a sweep over the number of layers for these MLPs and report performance of the best model.

\vspace{.3em}
\noindent\textbf{Model Components.} To probe the efficacy of the different components, we train ablated models
1) without visual features $v_i^j$ (`No CNN'),
2) without 3D coordinates $p_{i}$,
3) without the learnable camera embedding $k_j$ (`No LCE'),
4) replacing the attention based aggregation with mean pooling, and
5) removing the auxiliary loss term.

\Cref{table:ablation} reports the ablation results.
All models are trained on images from the RFF datasets downsampled to \qty{25}{\%}
of the original resolution (\numproduct{504x 378}).
We use the average metrics across all the scenes for comparison.

\subsection{Interpreting the model}
The use of transformers and epipolar geometry in our model permits interpretation of the results via the attention weights. We demonstrate this by extracting correspondences and depth maps.
Also, our use of a four-dimensional light field representation enables the construction of epipolar-plane images (EPI)~\cite{bolles1987epipolar}, which are interpretable reconstructions of the scene geometry.

\vspace{.3em}
\noindent\textbf{Dense correspondence.}
We can extract potential correspondences between a target ray
and a reference view $j$ by finding the largest attention weights in $\{\alpha_i^j\}_{1\le i \le P}$. 
\Cref{fig:correspondence} shows the weight distribution of putative correspondences over the epipolar line for four points of interest. For points (1) and (4) we observe unimodal distributions with peaks at the point of correspondence. For point (2) we notice some uncertainty, while for point (3), the distribution is multi-modal with peaks around each blade with the highest peak near the correct correspondence.
\begin{figure}[!t]
    \centering
    \begin{tabular}{c}
     \includegraphics[width=.45\textwidth]{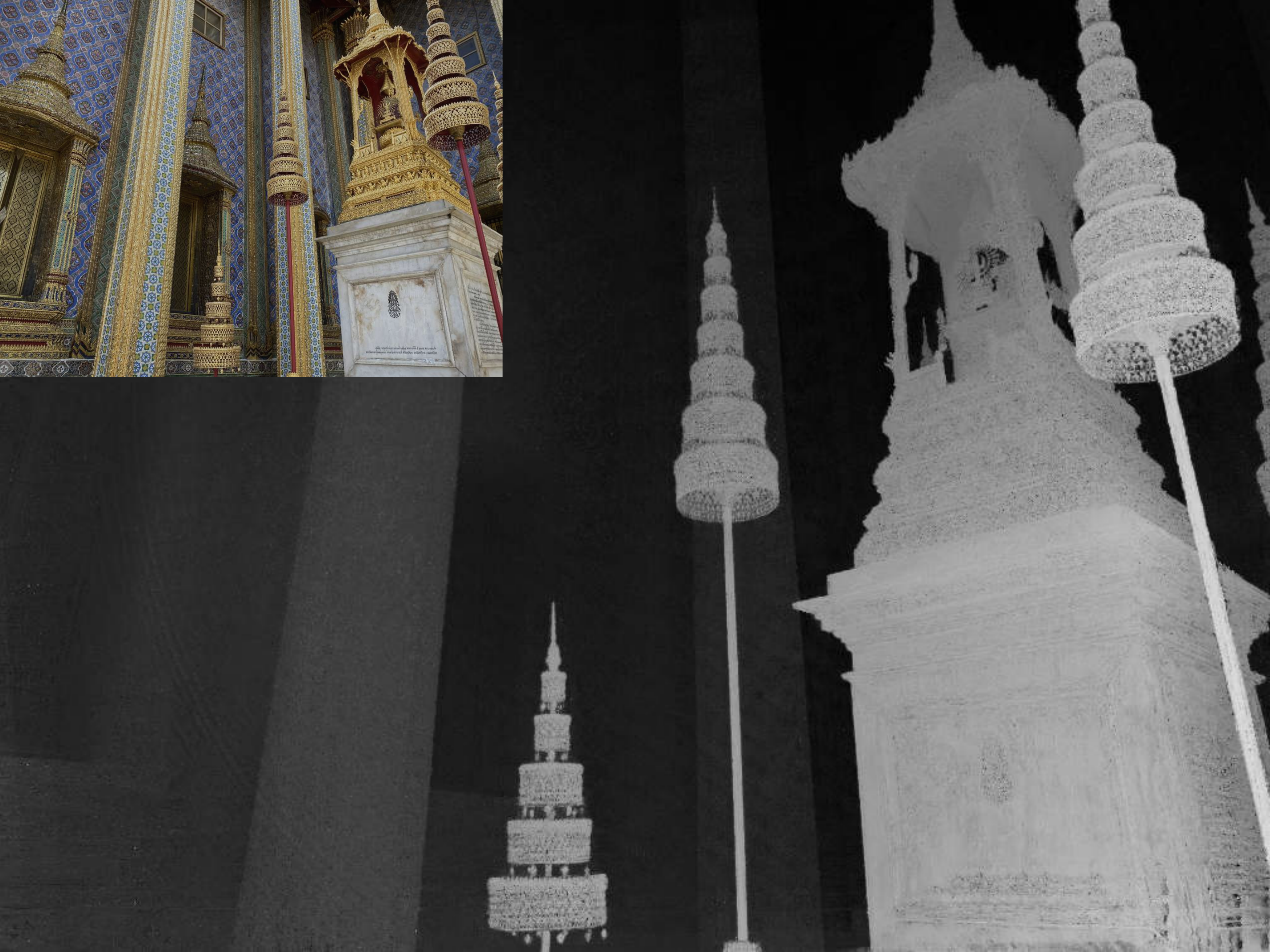} \\
    \end{tabular}
    \caption{\textbf{Disparity Map.}
      The per-point and per-view attention weights learned by our model can be applied
      to estimate a disparity map by aggregating
      the putative depths of each epipolar point on each reference view,
      for each target ray.
    }
    \label{fig:crest_disparity}
\end{figure}

\vspace{.3em}
\noindent\textbf{Disparity map.}
Since each epipolar point corresponds to the projection of the target ray at a certain depth
(the putative depth),
we can use the correspondence distributions to estimate a depth map for a target ray.
We first extract all the
epipolar point ($\alpha_i^j$) and reference view ($\beta^j$) attention weights
as described in \cref{sec:transformer},
then compute a weighted average of putative depths,
equivalent to applying \cref{eq:reg_color} with colors replaced by depths.
\Cref{fig:crest_disparity} shows an example of a the disparity map obtained for a test view of the \emph{Crest} scene from the Shiny dataset~\cite{wizadwongsa2021nex}.

\begin{figure}[!t]
    \centering
     \includegraphics[width=.47\textwidth]{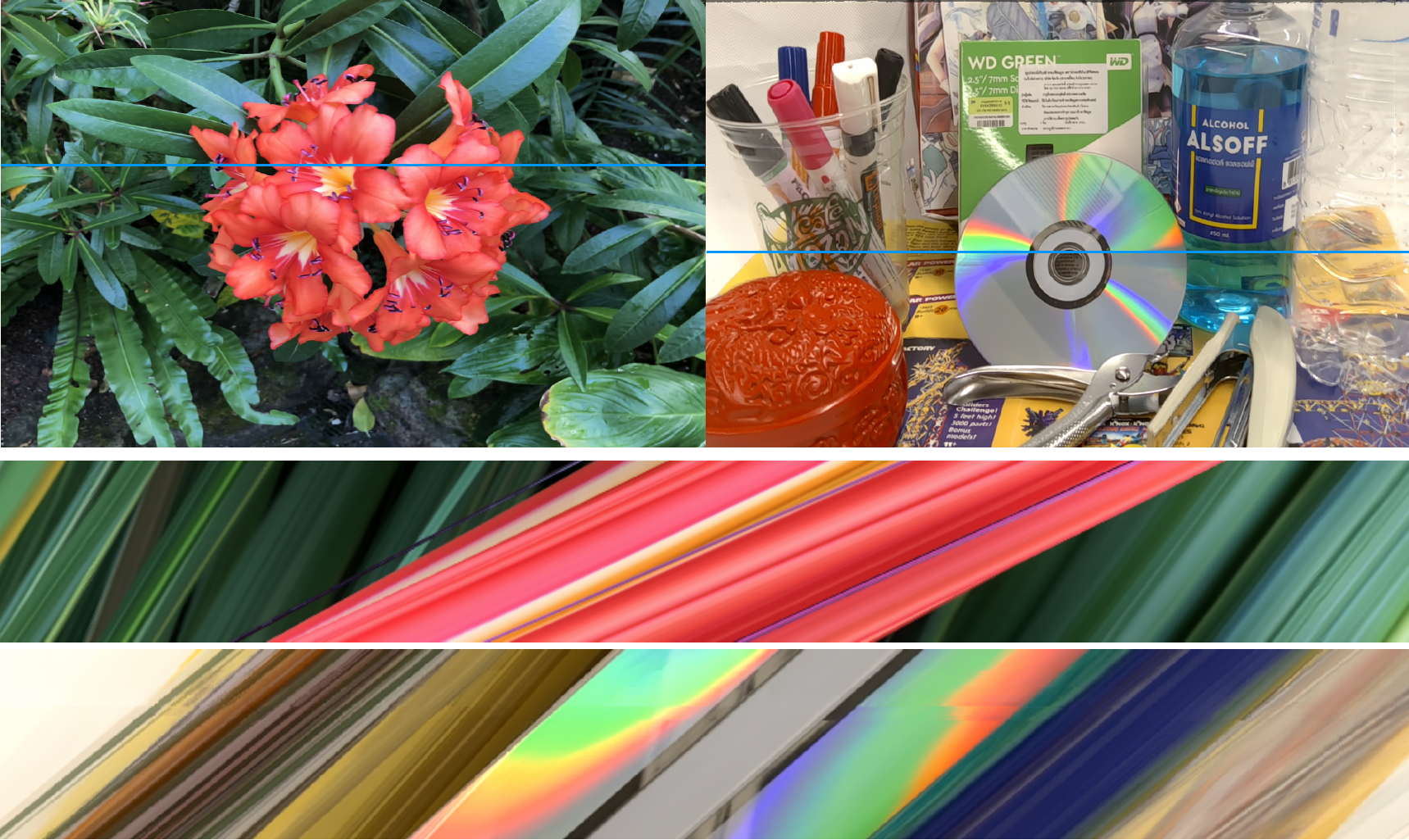} \\
     \caption{\textbf{Epipolar-plane images (EPI).}
       Our model represents the 4D light field, so constructing EPIs is natural.
       Each EPI vertically stacks images along the blue line,
       while the camera moves parallel to the blue line.
       Different depths show as lines of different slopes in the EPI,
       while view-dependent effects show as curves.
     }
    \label{fig:epi}
\end{figure}

\vspace{-1.2em}
\paragraph{Epipolar-plane images (EPI).}
For a 4D light slab representation,
we construct the EPI by querying our model with
two fixed and two variable coordinates, resulting in a 2D color image.
Physically, this corresponds to moving the camera along a 1D trajectory,
stacking the images of a line segment parallel to the trajectory.
EPIs encode information about specularities and scene geometry, where
diffuse points appear as lines and specular points appear as curves.
We show the epipolar slices for the \emph{CD} and \emph{Flower} scene in \cref{fig:epi}.
The \emph{Flower} scene is predominantly diffuse so we observe lines of varying slopes in the EPI,
with slopes inversely proportional to the depth.
For the  \emph{CD} scene, in addition to lines, we observe curves at regions corresponding to the interference pattern on the disk.
This is due to the change in virtual apparent depth of the specular points with change in view point~\cite{swaminathan2002motion,mildenhall2019local}.
\begin{figure}
    \centering
    \setlength\tabcolsep{1.3pt}
    \begin{tabular}{cc  cc}
        \shortstack{
            \includegraphics[width=0.11\textwidth]{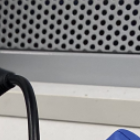} \\ Ground Truth
        } &
        \shortstack{
            \includegraphics[width=0.11\textwidth]{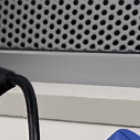} \\ \textbf{Ours}
        } &
        \shortstack{
            \includegraphics[width=0.11\textwidth]{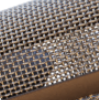} \\ Ground Truth
        } &
        \shortstack{
            \includegraphics[width=0.11\textwidth]{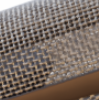} \\ \textbf{Ours}
        } 
    \end{tabular}
    \caption{\textbf{Failure Cases.}
      Our model is challenged by texture-less thin repeating structures.
      Left: It produces distorted circles on the grill-like structure in the \emph{Tools} scene from Shiny~\cite{wizadwongsa2021nex}.
      Right: Similar distortions appear for the \emph{Mic} scene from Blender~\cite{mildenhall2020nerf}.}
    \label{fig:limitation}
\end{figure}
\section{Limitations}
\label{sec:limitations}

Since our method relies on implicitly finding the correspondences for a target pixel in nearby views,  it is challenged by texture-less thin repeating structures. As shown in \cref{fig:limitation}, our model produces fuzzy details on the grill-like structure in the \emph{Tools} scene from Shiny and on the wire-mesh on the microphone from Blender.

Transformers are computationally expensive, resulting in slow training and inference times. Our method is around 8 times slower than Mip-NeRF~\cite{barron2021mip} on same hardware.
Our model does, however, compare favorably in terms of speed against other transformer-based models.
For example, NerFormer~\cite{reizenstein2021common} takes around \qty{180}{s} to render an \numproduct{800 x 800} image on a single V100 GPU, while our method takes  \qtyrange{60}{70}{\second} on the same hardware.
We notice that our model suffers from overhead of
1) random memory access on device and
2) data transfer between host and device.
We believe that it can be made more efficient with some engineering effort.

\section{Conclusion}
We present a light field based neural  rendering method for novel-view synthesis. Unlike prior volumetric rendering methods, our proposed model can naturally handle real-world illumination effects by learning the light field over a four-dimensional space. To address the dense sampling dependency of light field rendering, we introduced a two-stage framework that incorporates geometric inductive bias in the form of epipolar constraints. Our model leads to significant improvement over previous state-of-the-art model for view synthesis especially for scenes with challenging view-dependent effects. Finally, the design of our model allows extracting dense correspondences, disparity maps and epipolar-plane images without any additional training.

{\small
\setlength{\bibsep}{0pt}
\bibliographystyle{abbrvnat}
\bibliography{egbib}
}

\setcounter{page}{0}
\pagenumbering{arabic}
\setcounter{page}{1}

\setcounter{table}{0}
\renewcommand{\thetable}{\thesection.\arabic{table}}
\renewcommand{\thefigure}{\thesection.\arabic{figure}}
\appendix

\section{Additional implementation details}

\subsection{MLP architecture in ablation}
We detail the architecture of the `1-MLP` model  introduced in \cref{subsec:ablation}.
 We present the detailed architecture in \cref{table:architecture1}. We use a series of \texttt{DenseGeneral(DG)} layers and a final \texttt{Dense} available in Flax~\cite{flax2020github}. The architecture was determined by running a sweep over various depths. We found that further increase in model capacity leads to poor generalization.
\begin{table}[h]
\small
\centering
\begin{tabular}{@{}lll@{}}
\toprule
Layer          & Input Dimension & Output Dimension \\ \midrule
\texttt{DG}($F$, $256$)       & $B$ x $N$ x $P$ x $F$   & $B$ x $N$ x $P$ x $256$  \\
\texttt{DG\textsubscript{12}}($256$, $256$)   & $B$ x $N$ x $P$ x $256$   & $B$ x $N$ x $P$ x $256$  \\
\texttt{DG}($P$, $1$)        & $B$ x $N$ x $P$ x $256$   & $B$ x $N$ x $256$      \\
\texttt{DG}($N$, $1$)        & $B$ x $N$ x $256$         & $B$ x $256$          \\
\texttt{Dense}($256$, $1$)   & $B$ x $256$               & $B$ x $3$            \\ \bottomrule
\end{tabular}
\caption{\textbf{1-MLP Architecture.} We use notations $B$ for batch size, $N$ for number of reference views, $P$ for number of epipolar projection and $F$ for feature dimension. \texttt{DG\textsubscript{12}} represents 12 layers of \texttt{DenseGeneral}. We also add skip connections at every fourth layer in \texttt{DG\textsubscript{12}}. The final output corresponds to the predicted color. }
\label{table:architecture1}
\end{table}

\subsection{Spherical light field encoding}

For \ang{360} scenes, we use the two-sphere light field parametrization \cite{camahort1998uniformly}. Each ray is represented by two points on the sphere,
by the 4D tuple $(\theta_1, \phi_1, \theta_2, \phi_2)$.
To encode this representation we found advantageous to use the spherical harmonics
basis instead of the sinusoidals.
For a given ray, we evaluate a number of spherical harmonics at each intersection and
concatenate them to obtain its encoding,
\begin{equation}
    \tilde{Y}_{m}^{\ell}(\theta_1, \phi_1, \theta_2, \phi_2) = \left[ Y_{m}^{\ell}(\theta_1, \phi_1) \parallel Y_{m}^{\ell}(\theta_2, \phi_2) \right],
\end{equation}
where $Y_{m}^{\ell}(\theta, \phi)$ denotes the spherical harmonics of degree $\ell$ and order $m$ evaluated at $(\theta, \phi)$. 
In our experiments, we concatenate all the zonal and sectoral harmonics ($m=0$ and $m=\ell$) upto a maximum degree of $4$. %

To demonstrate the efficacy of the spherical harmonics encoding we conduct an ablation on the blender dataset where we replace the spherical encoding with the regular positional encoding in NeRF \cite{mildenhall2020nerf}. We refer to this model as Ours\textsubscript{P.E.}. We report the average metric on the blender dataset for this ablation in \cref{table:blender_sp_ablation}.

\begin{table}[!t]
\small
\centering
\begin{tabular}{@{}lcccc}
\toprule \toprule
Model                                        & PSNR $\uparrow$ & SSIM $\uparrow$ & LPIPS  $\downarrow$ & Avg.  $\downarrow$ \\ \midrule
Ours\textsubscript{P.E.}                                        & 33.18       & 0.979       & 0.027           & 0.0123         \\
Ours                                         & \bgd33.85       & \bgd0.981       & \bgd0.024           & \bgd0.0110         \\ \bottomrule \bottomrule
\end{tabular}
\caption{Light field encoding ablation on Blender dataset.}
\label{table:blender_sp_ablation}
\end{table}

\subsection{Metric computation}
To compute the SSIM metric we use the function available in scikit-image package. To compute the LPIPS on forward facing scene (RFF and Shiny), similar to NeX, we use the VGG model\footnotemark[2] from \cite{zhang2018perceptual}. On the blender scenes, similar to Mip-NeRF, we use the VGG model  available in tensorflow hub to compute LPIPS. We use two different implementation on LPIPS to ensure fairness of comparison.
\footnotetext[2]{We use the library provided by  \url{https://github.com/richzhang/PerceptualSimilarity}.}

\section{Additional results}

\subsection{Real-forward-facing dataset (RFF)}
The RFF dataset introduced by~\citet{mildenhall2019local} consists of 8 forward facing captures with each scene consisting of around 20 to 62 images.  We present the scene-wise breakdown of the results in \Cref{table:RFF_full}. The metrics for NeRF and NeX are the ones reported in NeX \cite{wizadwongsa2021nex}.

\begin{table*}[!ht]
\centering
\begin{tabular}{@{}lcccccccccccc}
\toprule
       &  & \multicolumn{3}{c}{PSNR} & & \multicolumn{3}{c}{SSIM} & & \multicolumn{3}{c}{LPIPS} \\ \cmidrule(l){3-5} \cmidrule(l){7-9}  \cmidrule(l){11-13} 
Model   & & NeRF       & NeX            & Ours       & & NeRF   & NeX          & Ours        & & NeRF    & NeX    & Ours   \\ \midrule
Fern&     & \bgl25.49  & \bgd25.63      & 24.86      & & 0.866  & \bgd0.887    & \bgl0.886   & & 0.278   & \bgl0.205  & \bgd0.135  \\
Flower&   & 27.54      & \bgl28.90      & \bgd29.82  & & 0.906  & \bgl0.933    & \bgd0.939   & & 0.212   & \bgl0.150  & \bgd0.107  \\
Fortress& & 31.34      & \bgl31.67      & \bgd33.22  & & 0.941  & \bgl0.952    & \bgd0.964   & & 0.166   & \bgl0.131  & \bgd0.119  \\
Horns&    & 28.02      & \bgl28.46      & \bgd29.78  & & 0.915  & \bgl0.934    & \bgd0.957   & & 0.258   & \bgl0.173  & \bgd0.121  \\
Leaves&   & 21.34      & \bgl21.96      & \bgd22.47  & & 0.782  & \bgl0.832    & \bgd0.856   & & 0.308   & \bgl0.173  & \bgd0.110  \\
Orchids&  & \bgl20.67  & 20.42          & \bgd21.05  & & 0.755  & \bgl0.765    & \bgd0.807   & & 0.312   & \bgl0.242  & \bgd0.173  \\
Room&     & 32.25      & \bgl32.32      & \bgd34.54  & & 0.972  & \bgl0.975    & \bgd0.987   & & 0.196   & \bgl0.161  & \bgd0.104  \\
Trex&     & 27.36      &\bgl 28.73      & \bgd30.34  & & 0.929  & \bgl0.953    & \bgd0.968   & & 0.234   & \bgl0.192  & \bgd0.143  \\ \midrule
\end{tabular}
\caption{Scene-wise breakdown of quantitative results on the Real Forward-Facing dataset.}
\label{table:RFF_full}
\end{table*}

\begin{table*}[!t]
\centering
\begin{tabular}{@{}lcccccccccccc}
\toprule
          &  & \multicolumn{3}{c}{PSNR} & & \multicolumn{3}{c}{SSIM} & & \multicolumn{3}{c}{LPIPS} \\ \cmidrule(l){3-5} \cmidrule(l){7-9}  \cmidrule(l){11-13}
Model   &  & NeRF        & NeX           & Ours    &      & NeRF   & NeX        & Ours &  & NeRF            & NeX        & Ours   \\ \midrule
CD  &      & 30.14       & \bgl31.43         & \bgd35.25    & & 0.093  & \bgl0.958  & \bgd0.989 & & 0.206   & \bgl0.129  & \bgd0.041  \\
Tools&     & \bgl27.45   & \bgd28.16     & 26.55        & & 0.938  & \bgd0.953  & \bgl0.945 & & 0.204   & \bgl0.151      & \bgd0.130  \\
Crest &    & 20.30       & \bgl21.23         & \bgd21.73    & & 0.670  & \bgl0.757  & \bgd0.797 & & 0.315   & \bgl0.162  & \bgd0.079  \\
Seasoning& & 27.79       & \bgd28.60     & \bgl28.34    & & 0.898  & \bgl0.928  & \bgd0.936 & & 0.276   & \bgl0.168      & \bgd0.102  \\
Food &     & \bgl23.32   & \bgd23.68     & 22.88        & & 0.796  & \bgd0.832  & \bgl0.821 & & 0.308   & \bgl0.203      & \bgd0.151  \\
Giants&    & 24.86       & \bgl26.00         & \bgd27.06    & & 0.844  & \bgl0.898  & \bgd0.928 & & 0.270   & \bgl0.147  & \bgd0.065  \\
Lab &      & 29.60       & \bgl30.43         & \bgd35.28    & & 0.936  & \bgl0.949  & \bgd0.989 & & 0.182   & \bgl0.146  & \bgd0.066  \\
Pasta &    & 21.23       & \bgd22.07     & \bgl21.63    & & 0.789  & \bgl0.844  & \bgd0.855 & & 0.311   & \bgl0.211      & \bgd0.096  \\ \bottomrule
\end{tabular}
\caption{Scene-wise breakdown of quantitative results on the Shiny dataset.}
\label{table:shiny_full}
\end{table*}
\subsection{Shiny dataset}
The Shiny dataset introduced in NeX~\cite{wizadwongsa2021nex} presents $8$ scenes with challenging view dependent effects, captured by forward-facing cameras. 
We present the scene-wise breakdown of the results in \cref{table:shiny_full}. The metrics for NeRF and NeX are the ones reported in NeX \cite{wizadwongsa2021nex}.

\subsection{Blender dataset}
The Blender dataset introduced by~\citet{mildenhall2020nerf} consists of 8 scenes each containing $800 \times 800$ resolution images rendered from viewpoints randomly sampled on a hemisphere around the object. We present the scene-wise breakdown of the results in \cref{table:blender_full}. The metrics for NeRF and Mip-NeRF were obtained from Mip-NeRF \cite{barron2021mip}.

\begin{table*}[]
\centering
\begin{tabular}{@{}lcccccccccccc}
\toprule
          &  & \multicolumn{3}{c}{PSNR} & & \multicolumn{3}{c}{SSIM} & & \multicolumn{3}{c}{LPIPS} \\ \cmidrule(l){3-5} \cmidrule(l){7-9}  \cmidrule(l){11-13} 
Model &    & NeRF  & Mip-NeRF & Ours & & NeRF  & Mip-NeRF & Ours   & & NeRF   & Mip-NeRF & Ours  \\ \midrule
Chair  &   & 34.08 & \bgl35.14    & \bgd35.30 & & 0.975 & \bgl0.981    & \bgd0.989 & & 0.026  & \bgl0.021    & \bgd0.012 \\
Drums  &   & 25.03 & \bgl25.48    & \bgd25.83 & & 0.925 & \bgl0.932    & \bgd0.955 & & 0.071  & \bgl0.065    & \bgd0.045 \\
Ficus  &   & 30.43 & \bgl33.29    & \bgd33.38 & & 0.967 & \bgl0.980    & \bgd0.987 & & 0.032  & \bgl0.020    & \bgd0.010 \\
Hotdog &   & 36.92 & \bgl37.48    & \bgd38.66 & & 0.979 & \bgl0.982    & \bgd0.993 & & 0.030  & \bgl0.027    & \bgd0.009 \\
Lego&      & 33.28 & \bgl35.7     & \bgd35.76 & & 0.968 & \bgl0.978    & \bgd0.989 & & 0.031  & \bgl0.021    & \bgd0.010 \\
Materials& & 29.91 & \bgl30.71    & \bgd35.10 & & 0.953 & \bgl0.959    & \bgd0.990 & & 0.047  & \bgl0.040    & \bgd0.011 \\
Mic &      & 34.53 & \bgd36.51    & \bgl35.32 & & 0.987 & \bgl0.991    & \bgd0.992 & & 0.012  & \bgl0.009    & \bgd0.008 \\
Ship &     & 29.36 & \bgl30.41    & \bgd30.94 & & 0.869 & \bgl0.882    & \bgd0.952 & & 0.150  & \bgl0.138    & \bgd0.084 \\ \bottomrule
\end{tabular}
\caption{Scene-wise breakdown of quantitative results on the Blender dataset.}
\label{table:blender_full}
\end{table*}

\section{Additional experiments and visualizations}

\subsection{Plücker coordinates}
Our experiments use ray parametrizations specific to the camera configuration of each type of scene.
For forward facing scenes, we employ the light slab light field representation,
while for \ang{360} scenes we use the two-sphere.
In this section, we explore the alternative of using Plücker coordinates,
which are generic and can represent any kind of camera configuration.
Since our architecture is agnostic to the light field parametrization,
we simply replace the input ray representation with the 6D Plücker coordinates to perform this experiment.
When using Plücker coordinates, we observe a drop of \qty{0.18}{\dB} PSNR on the RFF dataset as compared to the light slab representation.
Similarly, we observe a drop of \qty{0.25}{\dB} PSNR on the Blender dataset when replacing the two-sphere parametrization with Plücker coordinates.
This suggests that for particular configurations, the specific (and lower dimensional) ray parametrizations have a slight advantage over a generic parametrization such as Plücker coordinates.

\begin{figure}[!h]
\centering
\includegraphics[width=.4\textwidth]{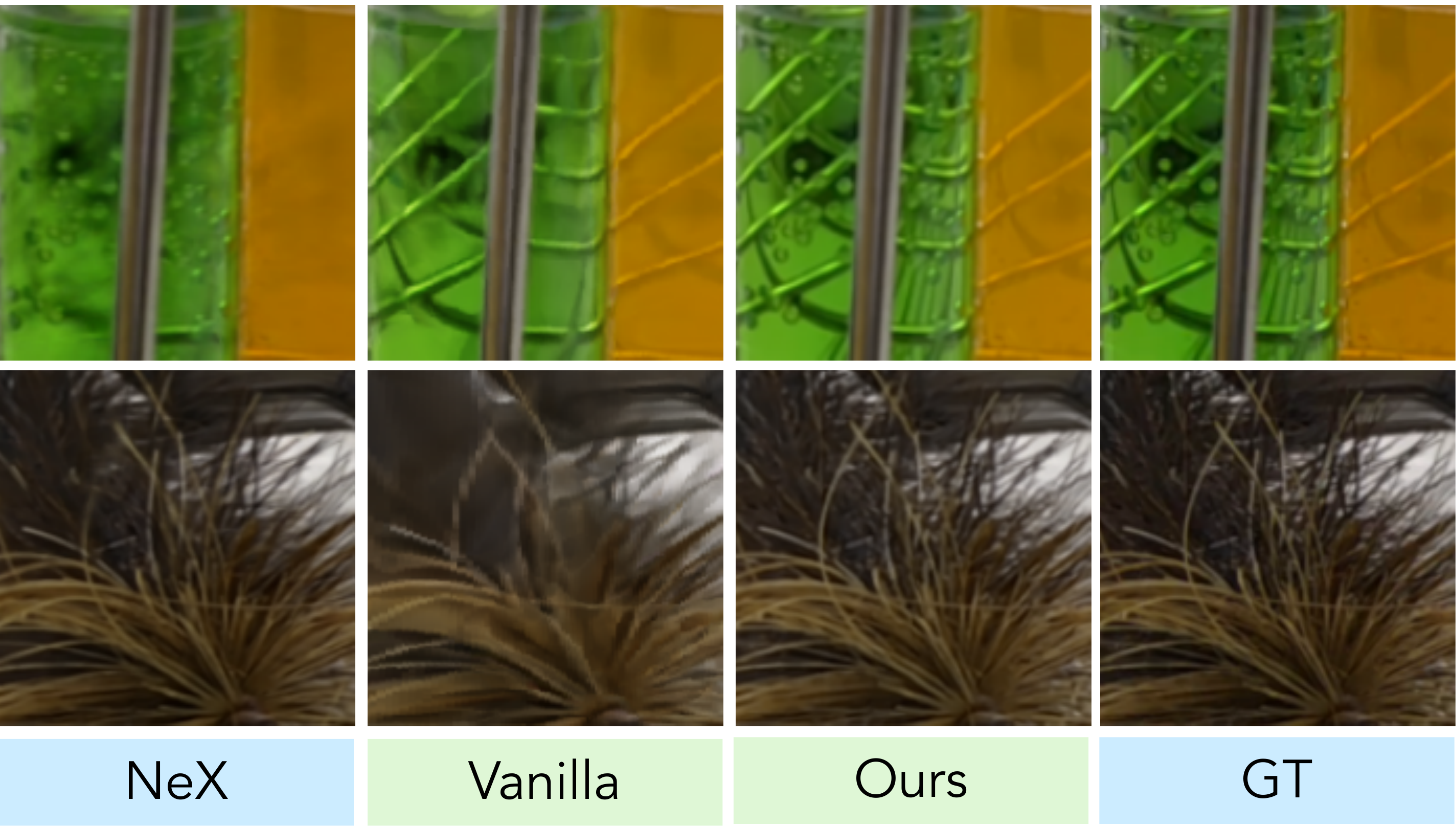}
\caption{Crops from two different regions of the \emph{Lab} scene in the Shiny dataset.
  The Vanilla-NLF model (described in Section 4.3) is able to retrieve a majority of the refraction details but fails to reproduce high frequencies.
  NeX reproduces sharp details but not the refractions. Our model does well in both regions.
}%
\label{fig:view_dependant}
\end{figure}
\subsection{Handling view-dependent effects}
While our model is built around geometric constraints (such as epipolar geometry), 
the attention-based modeling provides the capability to downweigh such constrains when not useful,
in order to more directly associate a color to ray coordinates (as in the Vanilla-NLF model described in Section 4.3).
We speculate that this, together with the convolutional features (which bring some context)
explains our superior performance on view-dependent effects. 

We run a mini-ablation to investigate this hypothesis.
\Cref{fig:view_dependant} shows, for the Lab scene from the Shiny dataset,
one crop with transparency/refraction and another that is diffuse and contains sharp details.
Our model works well on both regions which indicates its flexibility, in contrast with the baselines. 

\subsection{Visualizing view attention}
\begin{figure*}
    \centering
    \includegraphics[width=.95\textwidth]{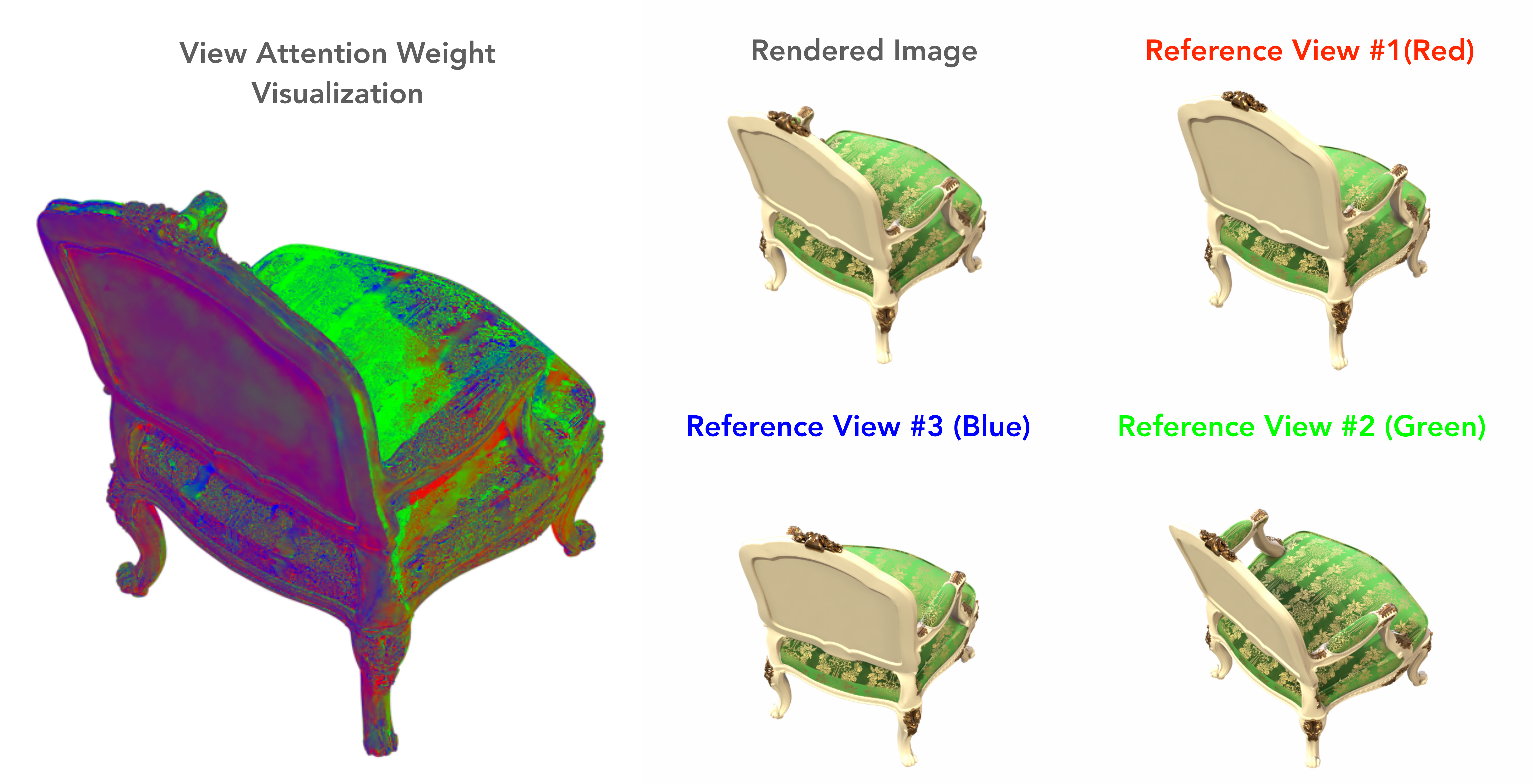}
    \caption{\textbf{View attention weight visualization.} We visualize the attention weights $\beta^{j}$ for each rendered pixel for a test image from the chair scene. For each target pixel, we consider three reference views. Thus we have three attention weights $\beta^{1}, \beta^{2} $ and $\beta^{3}$ corresponding to reference views 1, 2 and 3 respectively. We treat these attention weights as RGB value and visualize them as an image as shown above. Intuitively, this image shows the contribution of each reference view when rendering a pixel. For example, the cushion is predominantly green as it is most visible in second reference view. Similarly the back of the chair contains almost equal mix of red and blue as it is equally visible in reference views 1 and 3. We do not show the attention weights for the background pixels for clarity of visualization. }
    \label{fig:beta_viz}
\end{figure*}

The attention weights $\beta^{j}$ (in \cref{eq:neighbor_avg}) correspond to ``importance'' of each reference view when rendering a target pixel. We visualize these attention weight for a test image in the chair scene from Blender dataset in \cref{fig:beta_viz}. We explain the visualization process in the figure caption.

\end{document}